\documentclass[runningheads]{llncs}
\usepackage[T1]{fontenc}

\usepackage{graphicx,verbatim}
\usepackage{subcaption}
\usepackage{array}
\newcolumntype{C}[1]{>{\centering\arraybackslash}p{#1}}
\usepackage{multirow}
\usepackage{tabularx}

\begin{document}
\title{Diffusion-Based Synthesis of Complete 3D Biparametric Prostate MRI}
%

\author{Claudia Giardina\orcidID{0000-0002-2670-7069} \and
Montse Pardàs\orcidID{0000-0002-5861-6356} \and
Verónica Vilaplana\orcidID{0000-0001-6924-9961}}
\authorrunning{C. Giardina et al.}

\institute{Universitat Politècnica de Catalunya, Barcelona, Spain \\
\email{veronica.vilaplana@upc.edu}} 
  
\maketitle              
\begin{abstract}
Synthetic biparametric MRI (bpMRI) may help alleviate the limited availability of large and representative prostate MRI datasets for deep learning applications. In this work, we present a conditional 3D diffusion framework for anatomy-conditioned synthesis of complete prostate bpMRI from anatomical masks. The framework supports sequential synthesis, in which T2w and diffusion modalities are generated by separately trained models, and joint synthesis, in which all bpMRI modalities are generated simultaneously within a single model. Five configurations were trained using PI-CAI, FastMRI Prostate, and PROSTATEx datasets to investigate different conditioning schemes, target modalities, and synthesis strategies. Evaluation using distribution-based metrics, paired-image similarity, and qualitative assessment demonstrated the feasibility of both synthesis pathways. Joint synthesis achieved competitive distributional fidelity while generating complete bpMRI, whereas direct synthesis of ADC and computed high-b-value DWI outperformed their derivation from synthetic diffusion-weighted images.

\keywords{Diffusion models \and Prostate MRI \and Biparametric MRI \and Multimodal Image Synthesis \and Anatomical Conditioning}

\end{abstract}

\section{Introduction}

Prostate cancer is one of the most common malignancies in men, with more than 1.4 million new cases estimated worldwide in 2022~\cite{wcrf2022prostate}. Biparametric MRI (bpMRI) is central to the non-invasive assessment of clinically significant prostate cancer (csPCa) within the PI-RADS framework~\cite{turkbey2019pirads}. A bpMRI examination comprises T2-weighted imaging (T2w) and diffusion-weighted imaging (DWI) acquired at multiple b-values. From these acquisitions, quantitative apparent diffusion coefficient (ADC) maps can be derived, while very-high-b-value DWI may be acquired directly or computed from lower-b-value images.

BpMRI is increasingly used in deep learning applications such as lesion detection, segmentation, and csPCa classification~\cite{fouladi2026unet,chrysostomou2025multimodal,falana2023t2w}. However, model performance and generalization depend on the availability of large and representative annotated datasets~\cite{Bosma_2023}. This limitation has motivated increasing interest in prostate MRI synthesis as a means of augmenting existing datasets with synthetic images that reproduce clinically relevant imaging characteristics for downstream model development.

Early work on prostate MRI synthesis explored the generation of individual MRI modalities and paired diffusion images using 2D diffusion models~\cite{saeed2023biparametric}. More recently, several studies have extended prostate MRI synthesis to 3D volumes, including anatomy-conditioned T2w generation for segmentation augmentation~\cite{bashkanov2024zoneconditioned,giardinaT2wgen}, clinically conditioned T2w synthesis~\cite{grabke2025ccella}, and multimodal generation of T2w, high-b-value DWI, and ADC~\cite{grabke2025ccellaplus}.

Collectively, these studies demonstrate the feasibility of prostate MRI synthesis using diffusion models. However, anatomy-conditioned generation of complete 3D bpMRI from explicit prostate-zone and lesion masks remains largely unexplored. Furthermore, it remains unclear whether the diffusion component of bpMRI should be synthesized at the acquisition level by generating diffusion-weighted images and subsequently deriving ADC and computed high-b-value DWI, or directly at the biomarker level by generating ADC and computed high-b-value DWI themselves.

To investigate these alternative synthesis strategies, we present a conditional 3D diffusion framework for anatomy-conditioned generation of complete prostate bpMRI from anatomical masks. The framework supports both sequential synthesis, in which T2w and diffusion modalities are generated by separately trained models, and joint synthesis, in which all bpMRI modalities are generated simultaneously within a single model. It further supports both acquired diffusion targets (LBV and HBV) and diffusion-derived targets (ADC and cHBV), enabling the evaluation of alternative routes for complete bpMRI synthesis. The framework builds upon Med-DDPM~\cite{medddpm} and incorporates direct $x_0$ prediction, anisotropy-aware resolution changes, and flexible conditioning and target configurations. Five model configurations were trained using PI-CAI, FastMRI Prostate, and PROSTATEx data to evaluate different conditioning schemes, target modalities, and synthesis strategies for complete bpMRI generation.

\begin{figure}[t]
    \centering
    \includegraphics[width=\textwidth]{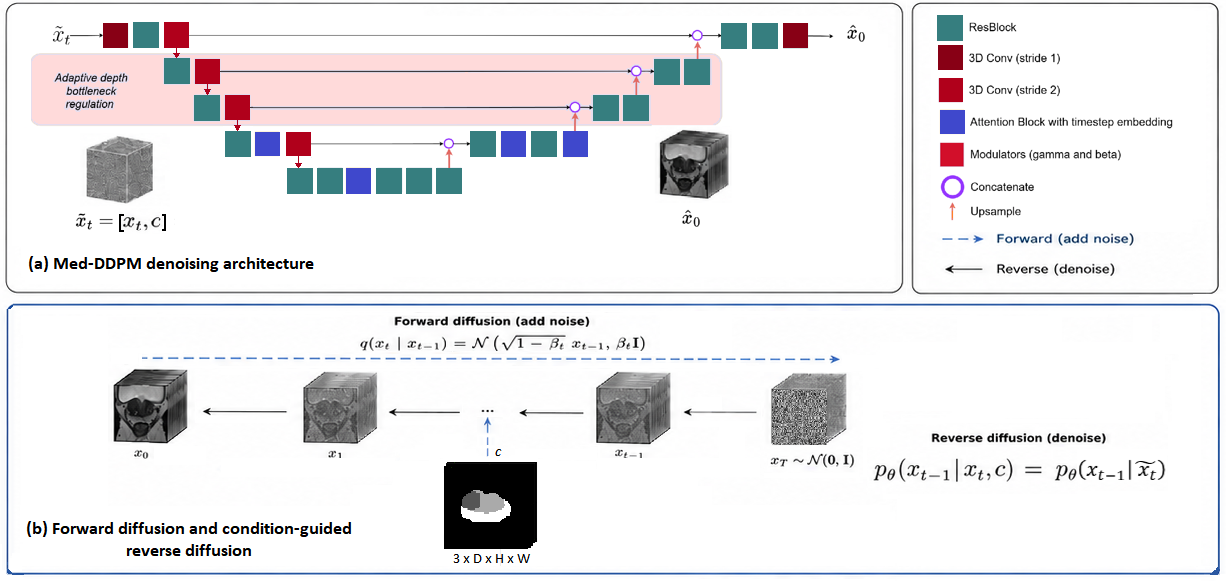}
    \caption{
Overview of the proposed framework for anatomy-conditioned prostate bpMRI synthesis. (a) Architecture of the conditional 3D diffusion model with anisotropy-aware depth regulation. (b) Example of the condition-guided reverse diffusion process for mask-conditioned T2w synthesis. The same formulation supports alternative conditioning and target configurations.
}
\label{fig:architecture}
\end{figure}

\section{Methodology}

\subsection{Framework Overview}

The proposed framework is based on a conditional 3D diffusion architecture derived from Med-DDPM~\cite{medddpm}. The framework employs a 3D U-Net denoiser in which the conditioning information $c$ is concatenated channel-wise with the noisy sample $x_t$ at each reverse-diffusion step, yielding the input $\tilde{x}_t=[x_t,c]$.

The framework introduces three modifications. First, the denoiser directly predicts the clean target volume, $\hat{x}_0=f_\theta([x_t,c],t)$, instead of the added noise. Preliminary T2w synthesis experiments indicated substantially improved distributional fidelity compared with noise prediction. This formulation was therefore adopted in all experiments.

Second, anisotropy-aware depth regulation (Fig.~\ref{fig:architecture}a) limits through-plane downsampling when the axial depth reaches a predefined minimum, preventing excessive compression of the comparatively low-resolution prostate MRI axis. The decoder mirrors this strategy during upsampling.

Third, the conditioning and output configurations extend beyond the original mask-conditioned formulation. The condition $c$ may consist of the anatomical mask alone (Fig.~\ref{fig:architecture}b) or of the mask concatenated with a T2w volume, while the target $x_0$ may contain either a single modality or multiple modalities. These configurations enable both sequential synthesis, in which T2w and diffusion targets are generated by separately trained models, and joint synthesis, in which all bpMRI modalities are generated simultaneously within a single model.

\subsection{Datasets and Image Preprocessing}

Three public datasets were used: PI-CAI~\cite{saha2024artificial}, comprising 1,500 bpMRI studies with T2w, ADC, and computed high-b-value DWI (cHBV); PROSTATEx~\cite{prostatex}, comprising 346 mpMRI studies; and FastMRI Prostate~\cite{tibrewala2024FastMRI}, comprising 312 bpMRI studies. PI-CAI provides prostate-zone and lesion annotations, while corresponding prostate and lesion masks were obtained for FastMRI and PROSTATEx. T2w, ADC, and cHBV were used from PI-CAI, whereas T2w, LBV, and HBV were used from FastMRI and PROSTATEx. FastMRI and PROSTATEx were preprocessed using QP-Prostate~\cite{quibim_qpprostate}, including image registration and quality control, yielding 235 and 281 usable studies, respectively. For PI-CAI, ADC and cHBV were registered to T2w using Elastix~\cite{klein2010elastix}, resulting in 1,158 usable studies. All volumes were resampled to $0.5\times0.5\times3.0$ mm, prostate-centred cropped to $192\times192\times20$ voxels, and intensity-normalized to $[-1,1]$.

\subsection{Experimental Design} \label{experiment_details}

In prostate bpMRI, DWI is acquired at multiple diffusion weightings, or $b$-values, typically including a low-$b$-value image (LBV, approximately $50$–$100~\mathrm{s/mm^2}$) and a high-$b$-value image (HBV, approximately $800$–$1000~\mathrm{s/mm^2}$). These acquisitions can be used to derive additional diffusion representations, including apparent diffusion coefficient (ADC) maps and computed high-$b$-value DWI (cHBV, approximately $1400~\mathrm{s/mm^2}$). This motivates two alternative diffusion synthesis strategies: generation of the acquired diffusion images (LBV and HBV), from which ADC and cHBV can subsequently be derived, and direct generation of the diffusion-derived representations ADC and cHBV.

Each diffusion target configuration was evaluated using both synthesis strategies supported by the proposed framework: sequential synthesis and joint synthesis. In sequential synthesis, a dedicated mask-conditioned model first generates T2w. A second model, conditioned on both the anatomical mask and the T2w volume, then generates either LBV/HBV or ADC/cHBV. During complete bpMRI synthesis, the generated T2w volume is used to condition the second-stage model. In joint synthesis, a single mask-conditioned model generates T2w together with either LBV/HBV or ADC/cHBV. The resulting experimental design comprises five model configurations, summarized in Table~\ref{tab:experimental_configurations}.

M1 was trained using PI-CAI and FastMRI (1615/120 train/test cases), M2 and M4 using PROSTATEx and FastMRI (476/40), and M3 and M5 using PI-CAI only (1146/12), according to the availability of the target modalities.

\begin{table}[t]
\centering
\caption{Experimental configurations for sequential and joint prostate bpMRI synthesis. M1--M5 denote the five trained model configurations. In sequential synthesis, models are listed in their order of application.}
\label{tab:experimental_configurations}
\footnotesize
\setlength{\tabcolsep}{4pt}
\renewcommand{\arraystretch}{1.0}
\begin{tabularx}{\linewidth}{@{}p{0.22\linewidth}p{0.16\linewidth}X@{}}
\hline
\textbf{Targets} &
\textbf{Strategy} &
\textbf{Condition $\rightarrow$ target(s)} \\
\hline
LBV, HBV &
Sequential &
M1: Mask $\rightarrow$ T2w \\
&
&
M2: Mask + T2w $\rightarrow$ LBV, HBV \\
&
Joint &
M4: Mask $\rightarrow$ T2w, LBV, HBV \\
\hline
ADC, cHBV &
Sequential &
M1: Mask $\rightarrow$ T2w \\
&
&
M3: Mask + T2w $\rightarrow$ ADC, cHBV \\
&
Joint &
M5: Mask $\rightarrow$ T2w, ADC, cHBV \\
\hline
\end{tabularx}
\end{table}

\subsection{Training Details}

All models were trained using Adam, scale-shift normalization, 250 diffusion steps, and the direct $x_0$-prediction objective. M1 and M2 used 64 initial U-Net channels (166.8M parameters), whereas M3–M5 used 96 channels (375.2M parameters). Learning rates were set to $2\times10^{-5}$ for M1, M4, and M5 and to $1\times10^{-5}$ for M2 and M3. All remaining optimization settings followed Med-DDPM~\cite{medddpm}.
To soften sharp boundaries in the anatomical condition, anisotropic Gaussian smoothing ($\sigma_z=0.4$, $\sigma_{xy}=1.0$) was applied stochastically to conditioning masks during training, with probabilities of 40\% for M1 and 50\% for M2–M5.
Training minimized a voxel-wise $L_1$ loss. For multi-target models, modality-specific losses were combined using empirically selected weights to balance learning across modalities.

\subsection{Evaluation}

\subsubsection{Distribution-based evaluation:}
The alignment between the real and synthetic image distributions was evaluated using FID~\cite{fid}, KID~\cite{kid}, and Precision and Recall~\cite{improved-Precision-Recall}. Metrics were computed on the 20 axial slices of each volume, treated as independent samples. The synthetic volumes used for this evaluation were generated from an augmented set of PI-CAI anatomical masks. Augmentation was designed to increase anatomical variability and balance lesion-positive and lesion-negative cases, and included prostate-level transformations (flipping and scaling) as well as lesion-level transformations (flipping, translation, and transfer between anatomically similar prostates). The resulting set comprised 1,500 masks. Synthetic volumes were generated using the corresponding sequential and joint synthesis pipelines. In sequential synthesis, T2w volumes were first generated from the augmented masks and subsequently used, together with the masks, to generate LBV/HBV or ADC/cHBV. In joint synthesis, all target modalities were generated directly from the augmented masks.
For each modality, synthetic volumes were compared with a size-matched real reference set: 1,500 real and synthetic PI-CAI T2w volumes, 1,158 real and synthetic PI-CAI ADC and cHBV volumes, and 516 real and synthetic FastMRI/PROSTATEx LBV and HBV volumes.

\subsubsection{Paired-image evaluation:}
PSNR and SSIM~\cite{ssim2004} were computed volume-wise between each generated volume and its matched real target for all test cases. Metrics were computed within the prostate region and averaged across datasets when applicable. 
For models conditioned only on anatomical masks (M1, M4, and M5), these metrics should be interpreted as measures of case-specific anatomical consistency rather than reconstruction fidelity.

\section{Results}

For T2w synthesis (Table~\ref{tab:t2w_distribution}), the joint models M4 and M5 achieved distribution-based performance comparable to that of the dedicated T2w model M1. M1 obtained the lowest FID and KID values, while both M4 and M5 achieved higher Precision and Recall. These results indicate that jointly generating T2w together with additional diffusion modalities does not substantially compromise the realism or diversity of the synthesized T2w images.

\begin{figure}[t]
    \centering
    \begin{minipage}[t]{0.45\linewidth}
        \centering
        \includegraphics[width=\linewidth]{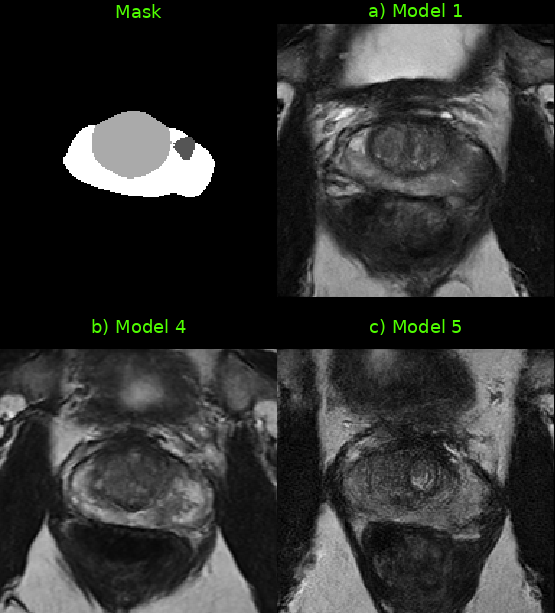}
    \end{minipage}
    \hspace{0.03\linewidth}
    \begin{minipage}[t]{0.45\linewidth}
        \centering
        \includegraphics[width=\linewidth]{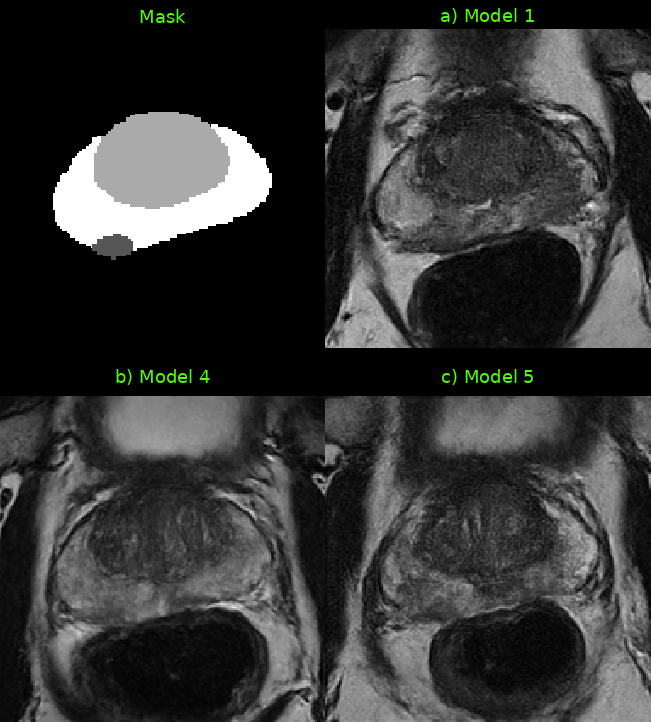}
    \end{minipage}
    \caption{Representative T2w synthesis examples from M1, M4, and M5 for two augmented masks.}
    \label{fig:t2w_model_comparison}
\end{figure}

The qualitative examples in Fig.~\ref{fig:t2w_model_comparison}
illustrate differences in the generated T2w appearances. M1 generally produced sharper anatomical boundaries and more detailed T2w texture, whereas the jointly generated T2w images from M4 and M5 appeared slightly smoother while remaining anatomically consistent with the conditioning masks.

The paired-image evaluation also favoured M4 (PSNR/SSIM: 20.19/0.30), followed by M1 (18.84/0.26) and M5 (18.69/0.23). Since all models were conditioned exclusively on anatomical masks, these metrics should be interpreted as measures of case-specific anatomical consistency rather than reconstruction fidelity. Overall, the results suggest that joint synthesis can generate realistic T2w images while preserving comparable case-specific agreement to a dedicated T2w model.

\begin{table}[t]
    \centering
    \caption{Distribution-based metrics for T2w synthesis.}
    \label{tab:t2w_distribution}
    \small
    \setlength{\tabcolsep}{6pt}
    \renewcommand{\arraystretch}{1.0}
    \begin{tabular}{lccc}
    \hline
    Metric & M1 & M4 & M5 \\
    \hline
    FID $\downarrow$ & \textbf{53.66} & 65.64 & 73.02 \\
    KID $\downarrow$ & \textbf{0.05} & 0.06 & 0.08 \\
    Precision $\uparrow$ & 0.41 & \textbf{0.65} & 0.59 \\
    Recall $\uparrow$ & 0.07 & \textbf{0.09} & \textbf{0.09} \\
    \hline
    \end{tabular}
    \end{table}

\begin{table}[t]
    \centering
    \caption{Distribution-based metrics for diffusion modality synthesis. For ADC and cHBV, the column ``M4-derived’’ corresponds to modalities computed from synthetic LBV and HBV generated by M4.}
    \label{tab:diffusion_distribution}
    \footnotesize
    \setlength{\tabcolsep}{2.5pt}
    \renewcommand{\arraystretch}{1.0}
    \begin{tabular}{lcc @{\hspace{8pt}} lcc}
    \hline
    \multicolumn{3}{c}{\textbf{LBV}} &
    \multicolumn{3}{c}{\textbf{HBV}} \\
    \cline{1-3}\cline{4-6}
    Metric & M2 & M4 &
    Metric & M2 & M4 \\
    \hline
    FID $\downarrow$ & 32.49 & \textbf{23.48} &
    FID $\downarrow$ & 37.57 & \textbf{25.50} \\
    KID $\downarrow$ & 0.021 & \textbf{0.016} &
    KID $\downarrow$ & 0.024 & \textbf{0.016} \\
    Prec. $\uparrow$ & 0.79 & \textbf{0.86} &
    Prec. $\uparrow$ & 0.46 & \textbf{0.67} \\
    Recall $\uparrow$ & 0.40 & \textbf{0.49} &
    Recall $\uparrow$ & \textbf{0.20} & 0.18 \\
    \hline
    \end{tabular}
    \vspace{2pt}
    \begin{tabular}{lccc @{\hspace{8pt}} lccc}
    \hline
    \multicolumn{4}{c}{\textbf{ADC}} &
    \multicolumn{4}{c}{\textbf{cHBV}} \\
    \cline{1-4}\cline{5-8}
    Metric & M3 & M5 & M4-derived &
    Metric & M3 & M5 & M4-derived \\
    \hline
    FID $\downarrow$ & 65.32 & \textbf{61.54} & 132.93 &
    FID $\downarrow$ & \textbf{42.60} & 51.23 & 76.88 \\
    KID $\downarrow$ & 0.069 & \textbf{0.064} & 0.136 &
    KID $\downarrow$ & \textbf{0.038} & 0.043 & 0.061 \\
    Prec. $\uparrow$ & \textbf{0.75} & \textbf{0.75} & 0.42 &
    Prec. $\uparrow$ & 0.69 & 0.69 & \textbf{0.70} \\
    Recall $\uparrow$ & \textbf{0.29} & \textbf{0.29} & 0.02 &
    Recall $\uparrow$ & \textbf{0.35} & 0.31 & 0.03 \\
    \hline
    \end{tabular}
    \end{table}

\begin{figure}[t]
\centering
\begin{minipage}[t]{0.42\textwidth}
\centering
\includegraphics[width=\textwidth]{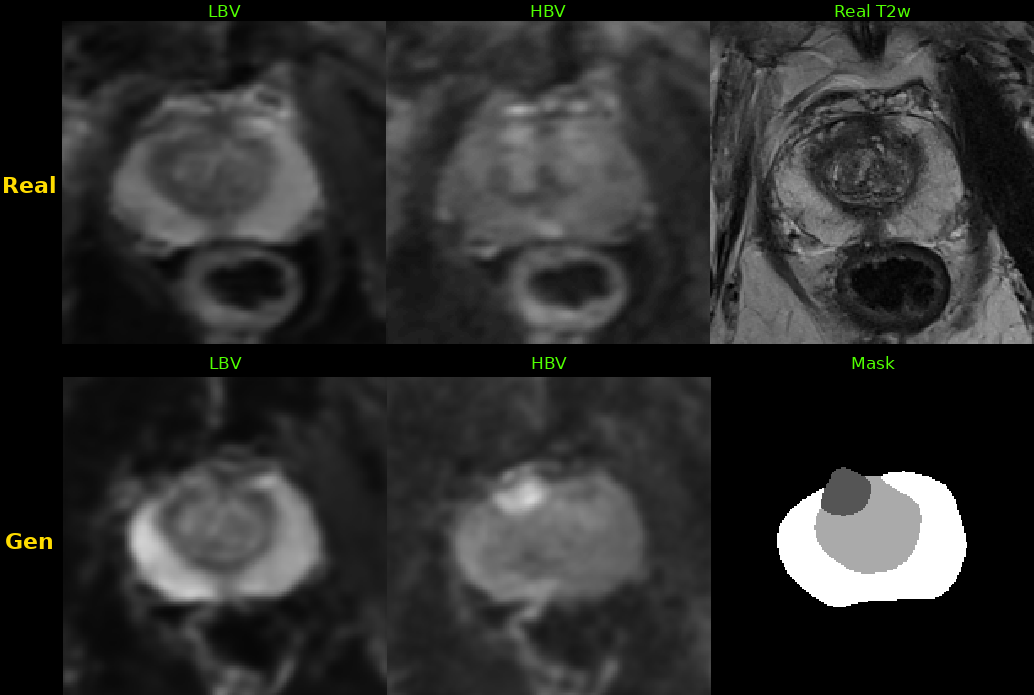}
\end{minipage}
\hfill
\begin{minipage}[t]{0.55\textwidth}
\centering
\includegraphics[width=\textwidth]{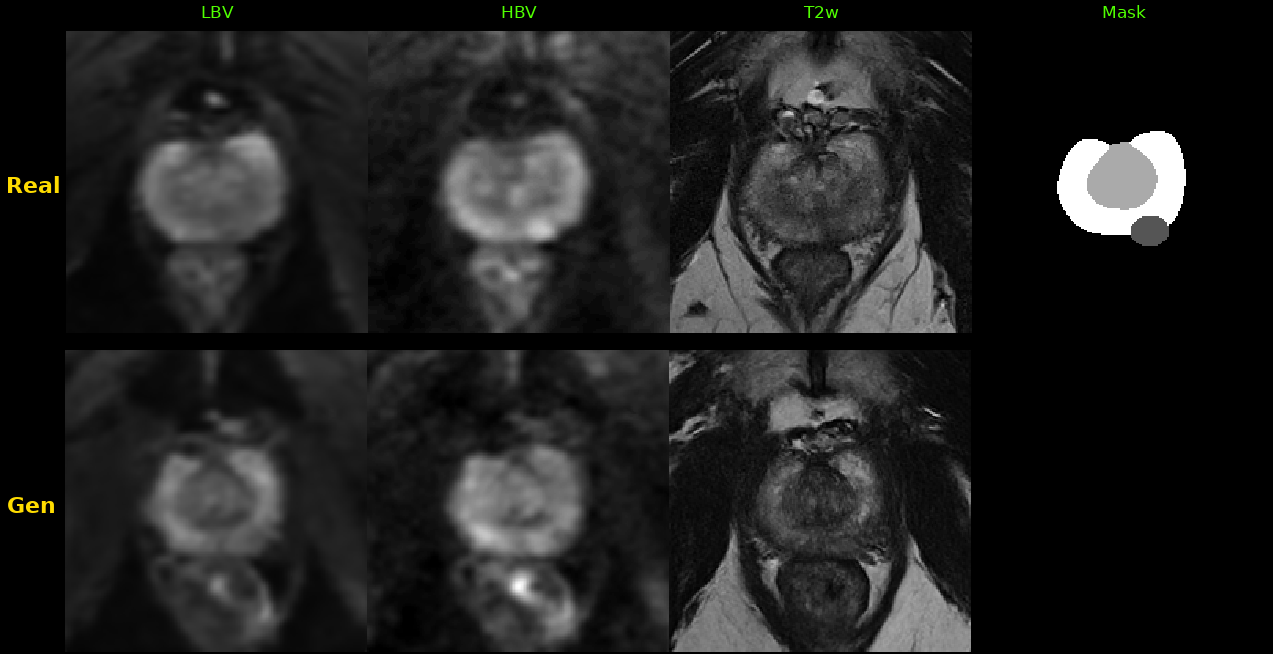}
\end{minipage}
\caption{Representative LBV/HBV synthesis examples from the sequential model M2 (left) and the joint model M4 (right).}
\label{fig:lbv_hbv_qualitative}
\end{figure}

For LBV and HBV synthesis (Table~\ref{tab:diffusion_distribution}),
the joint model M4 achieved better distribution-based performance
than the sequential model M2, with consistently lower FID and KID
values and higher Precision, while Recall was higher for LBV and
slightly lower for HBV. These results indicate that jointly
generating T2w, LBV, and HBV can reproduce the distribution and
variability of real diffusion-weighted images without degrading
image realism.

The paired-image evaluation favoured M2, which achieved higher LBV SSIM (0.72 vs.\ 0.63) and higher HBV PSNR/SSIM (17.11/0.57 vs.\ 15.74/0.53), while M4 obtained a slightly higher LBV PSNR (20.41 vs.\ 20.14). This difference is consistent with the additional real T2w conditioning available to M2. Overall, joint synthesis better matched the real image distribution, whereas sequential synthesis preserved stronger case-specific correspondence.

The examples in Fig.~\ref{fig:lbv_hbv_qualitative} show that both models reproduced the expected contrast differences between LBV and HBV while preserving prostate morphology. In some cases, M2 produced lesion regions with sharper, mask-conforming boundaries than those observed in real DWI, suggesting a degree of mask imprinting. This observation is qualitative and was not assessed by experts.

\begin{figure}[t]
\centering
\begin{minipage}[t]{0.41\textwidth}
\centering
\includegraphics[width=\textwidth]{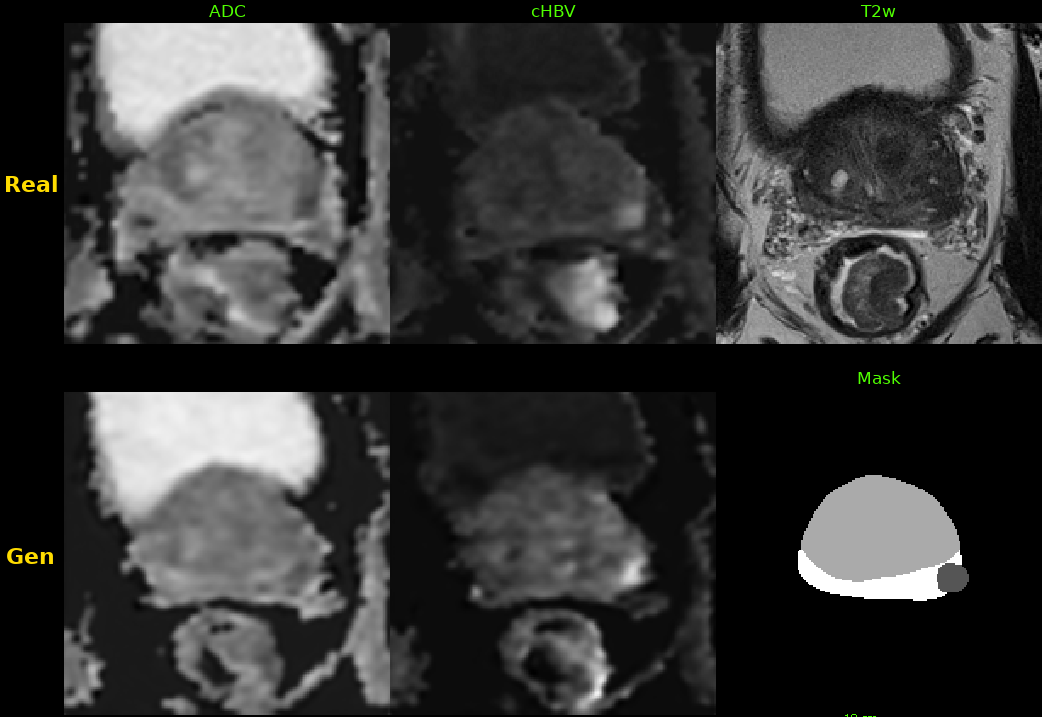}
\end{minipage}
\hfill
\begin{minipage}[t]{0.56\textwidth}
\centering
\includegraphics[width=\textwidth]{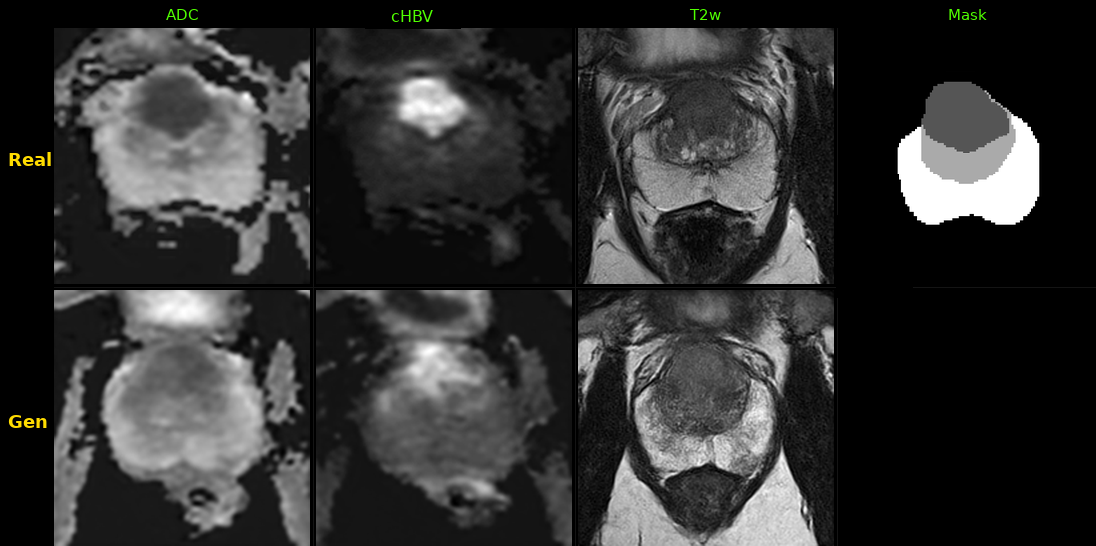}
\end{minipage}
\caption{Representative ADC/cHBV synthesis examples from the sequential model M3 (left) and the joint model M5 (right).}
\label{fig:adc_cHBV_qualitative}
\end{figure}

For ADC and cHBV synthesis (Table~\ref{tab:diffusion_distribution}), the sequential model M3 and the joint model M5 achieved very similar distribution-based performance. ADC metrics were nearly identical, while M3 showed a slight advantage for cHBV, achieving lower FID and KID values and higher Recall. These results indicate that jointly generating T2w, ADC, and cHBV does not substantially compromise the quality of the diffusion-derived modalities.

A clearer difference emerged when comparing direct synthesis with
the acquired-DWI pathway. ADC and cHBV computed from synthetic LBV
and HBV generated by M4 showed substantially worse distribution-based
performance than the directly synthesized modalities produced by
M3 and M5. This degradation was observed across all metrics except
cHBV Precision, which was slightly higher for the derived images,
suggesting that errors introduced during LBV and HBV synthesis
propagate to the derived ADC and cHBV representations.

The paired-image evaluation also favoured M3, which achieved higher PSNR/ SSIM values for both ADC (20.34/0.59 vs.\ 19.78/0.52) and cHBV (16.01/0.38 vs.\ 14.60/0.35). As in the LBV/HBV experiments, this advantage should be interpreted in light of the additional real T2w conditioning available to M3.

The examples in Fig.~\ref{fig:adc_cHBV_qualitative} show that both models reproduced the expected diffusion contrast patterns and lesion-associated signal changes while preserving consistency with the conditioning anatomy.

\begin{figure}[t]
\centering
\includegraphics[width=0.85\linewidth]
{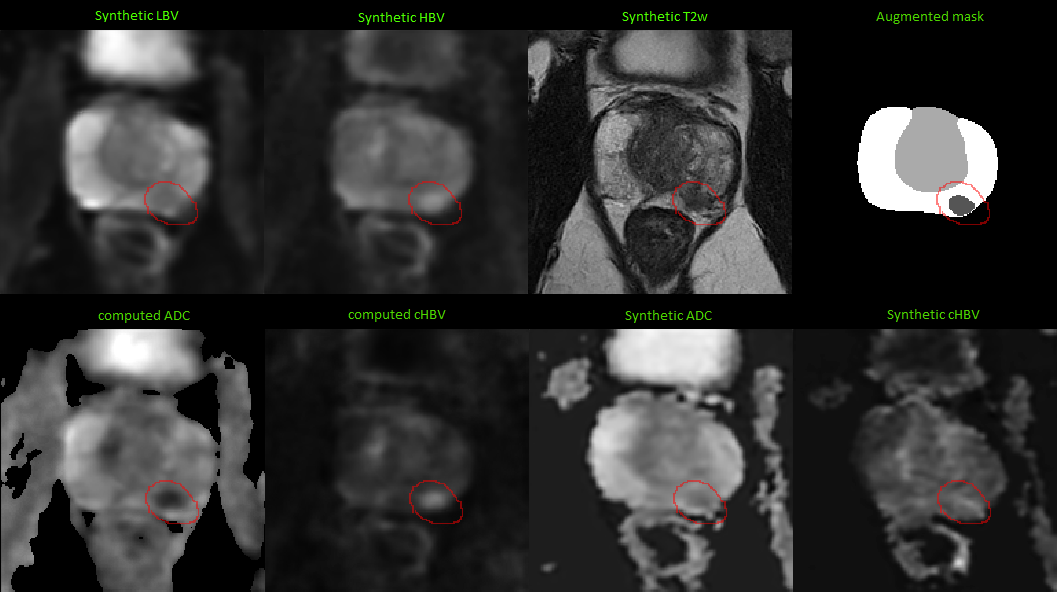}
\caption{Comparison of derived (left) and directly generated (right) ADC and cHBV. The lesion is highlighted by the red circle.
}
\label{fig:computed_vs_direct_diffusion}
\end{figure}

Figure~\ref{fig:computed_vs_direct_diffusion} illustrates the two investigated routes for obtaining ADC and cHBV: derivation from synthetic LBV/HBV and direct synthesis. Consistent with the quantitative results, the directly generated modalities exhibit patterns that more closely resemble those of the real data, whereas artefacts introduced during LBV and HBV synthesis may propagate to the derived ADC and cHBV images.

\section{Conclusion}

This study presented a conditional 3D DDPM framework for anatomy-conditioned synthesis of prostate bpMRI, supporting both sequential and joint generation of T2w and diffusion modalities from anatomical masks. The results demonstrate the feasibility of both synthesis strategies. Joint synthesis showed modality-specific differences in distributional
fidelity relative to dedicated models while generating complete bpMRI
examinations within a single model. For diffusion-derived modalities, direct synthesis of ADC and cHBV consistently achieved better distribution-based performance than derivation from synthetic LBV and HBV, suggesting that errors introduced during diffusion-image synthesis may propagate to the derived representations.

Although the quantitative evaluation and qualitative examples support the plausibility of the generated images, further validation is required to determine whether diagnostically relevant characteristics are preserved and whether the generated data provide clinical value. Future work will therefore include expert radiological assessment and evaluation in downstream tasks such as csPCa classification, lesion detection, and prostate segmentation.

\begin{credits}
\subsubsection{\ackname} This work was supported by the European project Federated Learning and mUlti-party computation Techniques for prostatE cancer (HORIZON-101095382-FLUTE), the Spanish Research Agency (AEI) under project PID 2023-148614OB-I00 funded by  MICIU/AEI /10.13039/501100011033 and by FEDER, EU, and the FPI-Ministerio PRE-2021-098481 grant.

\subsubsection{\discintname}
 The authors declare no competing interests.
\end{credits}

\bibliographystyle{splncs04}
\bibliography{bibliography}

@ARTICLE{medddpm,
    author={Dorjsembe, Zolnamar and Pao, Hsing-Kuo and Odonchimed, Sodtavilan and Xiao, Furen},
    journal={IEEE Journal of Biomedical and Health Informatics},
    title={Conditional Diffusion Models for Semantic 3D Brain MRI Synthesis},
    year={2024},
    volume={28},
    number={7},
    pages={4084-4093},
    doi={10.1109/JBHI.2024.3385504}
}

@article{saha2024artificial,
  title={Artificial intelligence and radiologists in prostate cancer detection on MRI (PI-CAI): an international, paired, non-inferiority, confirmatory study},
  author={Saha, Anindo and Bosma, Joeran S and Twilt, Jasper J and van Ginneken, Bram and Bjartell, Anders and Padhani, Anwar R and Bonekamp, David and Villeirs, Geert and Salomon, Georg and Giannarini, Gianluca and others},
  journal={The Lancet Oncology},
  volume={25},
  number={7},
  pages={879--887},
  year={2024},
  publisher={Elsevier}
}

@article{tibrewala2024fastmri,
  title={FastMRI Prostate: A public, biparametric MRI dataset to advance machine learning for prostate cancer imaging},
  author={Tibrewala, Radhika and Dutt, Tarun and Tong, Angela and Ginocchio, Luke and Lattanzi, Riccardo and Keerthivasan, Mahesh B. and Baete, Steven H. and Chopra, Sumit and Lui, Yvonne W. and Sodickson, Daniel K. and Chandarana, Hersh and Johnson, Patricia M.},
  journal={Scientific Data},
  volume={11},
  number={1},
  pages={404},
  year={2024},
  doi={10.1038/s41597-024-03252-w}
}

@misc{prostatex,
    author = {Litjens, G. and Debats, O. and Barentsz, J. and
              Karssemeijer, N. and Huisman, H.},
    title  = {{SPIE-AAPM-NCI PROSTATEx} Challenges ({PROSTATEx}):
              The Cancer Imaging Archive},
    url    = {https://wiki.cancerimagingarchive.net/pages/viewpage.action?pageId=23691656},
    doi    = {10.7937/K9TCIA.2017.MURS5CL},
    note   = {Online; accessed September 22, 2023}
}

@article{klein2010elastix,
  title={Elastix: a toolbox for intensity-based medical image registration},
  author={Klein, Stefan and Staring, Marius and Murphy, Keelin and Viergever, Max A and Pluim, Josien PW},
  journal={IEEE transactions on medical imaging},
  volume={29},
  number={1},
  pages={196--205},
  year={2010},
  publisher={IEEE}
}

@misc{quibim_qpprostate,
  author       = {{Quibim}},
  title        = {{QP-Prostate}: AI for prostate cancer detection and diagnosis},
  howpublished = {\url{https://quibim.ai/qp-prostate/}},
  note         = {Accessed: 2026-05-28}
}

@article{fid,
  author       = {Martin Heusel and
                  Hubert Ramsauer and
                  Thomas Unterthiner and
                  Bernhard Nessler and
                  G{\"{u}}nter Klambauer and
                  Sepp Hochreiter},
  title        = {GANs Trained by a Two Time-Scale Update Rule Converge to a Nash Equilibrium},
  journal      = {CoRR},
  volume       = {abs/1706.08500},
  year         = {2017},
  url          = {http://arxiv.org/abs/1706.08500},
  eprinttype    = {arXiv},
  eprint       = {1706.08500},
  bibsource    = {dblp computer science bibliography, https://dblp.org}
}

@misc{kid,
      title={Demystifying MMD GANs}, 
      author={Mikołaj Bińkowski and Danica J. Sutherland and Michael Arbel and Arthur Gretton},
      year={2021},
      eprint={1801.01401},
      archivePrefix={arXiv},
      primaryClass={stat.ML}
}

@article{ssim2004,
  title={Image quality assessment: From error visibility to structural similarity},
  author={Wang, Zhou and Bovik, Alan C and Sheikh, Hamid R and Simoncelli, Eero P},
  journal={IEEE Transactions on Image Processing},
  volume={13},
  number={4},
  pages={600--612},
  year={2004},
  publisher={IEEE},
  doi={10.1109/TIP.2003.819861}
}

@misc{improved-precision-recall,
      title={Improved Precision and Recall Metric for Assessing Generative Models}, 
      author={Tuomas Kynkäänniemi and Tero Karras and Samuli Laine and Jaakko Lehtinen and Timo Aila},
      year={2019},
      eprint={1904.06991},
      archivePrefix={arXiv},
      primaryClass={stat.ML}
}

@misc{wcrf2022prostate,
  author       = {{World Cancer Research Fund International}},
  title        = {Worldwide cancer data: Global cancer incidence in men, 2022},
  howpublished = {\url{https://www.wcrf.org/preventing-cancer/cancer-statistics/worldwide-cancer-data/}},
  note         = {Accessed: 2026-06-09}
}

@misc{grabke2025ccellaplus,
      title={Mitigating 3D Prostate Biparametric MRI Data Scarcity through Domain Adaptation using Locally-Trained Latent Diffusion Models for Prostate Cancer Detection}, 
      author={Emerson P. Grabke and Babak Taati and Masoom A. Haider},
      year={2026},
      eprint={2507.06384},
      archivePrefix={arXiv},
      primaryClass={eess.IV},
      url={https://arxiv.org/abs/2507.06384}, 
}

@InProceedings{giardinaT2wgen,
author="Giardina, Claudia
and Vila-Bagaria, Sigrid
and Alborn{\'e}s, Carlos
and Pard{\`a}s, Montse
and Vilaplana, Ver{\'o}nica",
editor="Mukhopadhyay, Anirban
and Oksuz, Ilkay
and Engelhardt, Sandy
and Mehrof, Dorit
and Yuan, Yixuan",
title="Anatomically Guided 3D Diffusion Models for Synthetic Prostate MRI Generation",
booktitle="Deep Generative Models",
year="2026",
publisher="Springer Nature Switzerland",
address="Cham",
pages="288--297",
isbn="978-3-032-05472-2"
}

@article{turkbey2019pirads,
  title={Prostate Imaging Reporting and Data System Version 2.1: 2019 Update of Prostate Imaging Reporting and Data System Version 2},
  author={Turkbey, Baris and Rosenkrantz, Andrew B. and Haider, Masoom A. and Padhani, Anwar R. and Villeirs, Geert and Macura, Katarzyna J. and Tempany, Clare M. and Choyke, Peter L. and Cornud, Francois and Margolis, Daniel J. A. and others},
  journal={European Urology},
  volume={76},
  number={3},
  pages={340--351},
  year={2019},
  doi={10.1016/j.eururo.2019.02.033}
}

@inproceedings{chrysostomou2025multimodal,
  author    = {Chrysostomou, Charalambos and Dovrolis, Constantinos},
  title     = {Multi-modal Deep Learning with Spatial Transformers for Biparametric MRI Prostate Cancer Classification},
  booktitle = {Artificial Intelligence in Healthcare: Second International Conference, AIiH 2025, Cambridge, UK, September 8--10, 2025, Proceedings, Part I},
  year      = {2025},
  pages     = {408--421},
  publisher = {Springer},
  doi       = {10.1007/978-3-032-00652-3_29},
  isbn      = {978-3-032-00651-6}
}

@article{fouladi2026unet,
  title={Exploring UNet-based models for prostate lesion segmentation from multi-sequence MRI (T2W, ADC, DWI)},
  author={Fouladi, Saman and Darvizeh, Fatemeh and Gianini, Gabriele and Di Meo, Rosario and Di Palma, Luca and Damiani, Ernesto and Maiocchi, Alessandro and Fazzini, Deborah and Al{\`i}, Marco},
  journal={World Wide Web},
  volume={29},
  number={4},
  year={2026},
  doi={10.1007/s11280-025-01392-6}
}

@inproceedings{falana2023t2w,
  title={Clinically Significant Prostate Cancer Classification Using Anatomical T2W MRI Sequences and Deep Learning},
  author={Falana, William Olurotimi and Serener, Ali and Serte, Sertan},
  booktitle={2023 7th International Symposium on Multidisciplinary Studies and Innovative Technologies (ISMSIT)},
  pages={1--6},
  year={2023},
  doi={10.1109/ISMSIT58785.2023.10304897}
}

@article{Bosma_2023,
   title={Semisupervised Learning with Report-guided Pseudo Labels for Deep                     Learning–based Prostate Cancer Detection Using Biparametric                     MRI},
   volume={5},
   ISSN={2638-6100},
   url={http://dx.doi.org/10.1148/ryai.230031},
   DOI={10.1148/ryai.230031},
   number={5},
   journal={Radiology: Artificial Intelligence},
   publisher={Radiological Society of North America (RSNA)},
   author={Bosma, Joeran S. and Saha, Anindo and Hosseinzadeh, Matin and Slootweg, Ivan and de Rooij, Maarten and Huisman, Henkjan},
   year={2023},
   month=sep }

@inproceedings{bashkanov2024zoneconditioned,
  title={Augmenting Prostate MRI Dataset with Synthetic Volumetric Images from Zone-Conditioned Diffusion Generative Model},
  author={Bashkanov, Oleksii and Rak, Marko and Engelage, Lucas and Hansen, Christian},
  booktitle={MICCAI Workshop on Deep Generative Models},
  pages={160--168},
  year={2024},
  organization={Springer}
}

@misc{saeed2023biparametric,
      title={Bi-parametric prostate MR image synthesis using pathology and sequence-conditioned stable diffusion}, 
      author={Shaheer U. Saeed and Tom Syer and Wen Yan and Qianye Yang and Mark Emberton and Shonit Punwani and Matthew J. Clarkson and Dean C. Barratt and Yipeng Hu},
      year={2023},
      eprint={2303.02094},
      archivePrefix={arXiv},
      primaryClass={eess.IV},
      url={https://arxiv.org/abs/2303.02094}, 
}

@misc{grabke2025ccella,
title={Prompt-Guided Latent Diffusion with Predictive Class Conditioning for 3D Prostate MRI Generation},
author={Emerson P. Grabke and Masoom A. Haider and Babak Taati},
year={2025},
eprint={2506.10230},
archivePrefix={arXiv},
primaryClass={eess.IV},
url={https://arxiv.org/abs/2506.10230},
}
\end{document}